%% file: compressed_cnn.tex
\documentclass{article}
 \usepackage[T1]{fontenc}
\usepackage[flushleft]{threeparttable}
\usepackage{algorithm}
\usepackage{algorithmic}
\usepackage{epsfig}
\usepackage{amssymb}
\usepackage{amsthm}
\usepackage{amsmath}
\usepackage{adjustbox}
\usepackage{dsfont}
\usepackage{bm}
\usepackage{amsfonts}
\usepackage{array}
\usepackage{subfigmat}
\usepackage{color}
\usepackage{booktabs}
\usepackage{url}
\usepackage[colorlinks=true,linkcolor=blue,citecolor=blue]{hyperref}
\usepackage{stmaryrd}
\usepackage{graphicx}
\usepackage{caption}
\usepackage{natbib}
\usepackage{float}

\DeclareMathOperator*{\E}{\mathbb{E}}

\begin{document}

 \begin{center}
 \Large
\textbf{A priori compression of convolutional neural networks for wave simulators \\}

\hspace{8pt}

\large
Hamza BOUKRAICHI$^{1,2}$, Nissrine AKKARI$^1$, \\ Fabien CASENAVE$^1$,  David RYCKELYNCK$^2$ \\

\hspace{8pt}

\small  
$^1$ Safran Tech \\
Etablissement Paris Saclay \\
Rue des Jeunes Bois-Chateaufort, 78114 Magny-Les-Hameaux, France\\ ~~ \\
$^2$ MINES ParisTech, PSL University \\
MAT - Centre des matériaux \\
CNRS UMR 7633, BP 87 91003 Evry, France
\end{center}

\begin{abstract}

Convolutional neural networks are now seeing widespread use in a variety of fields, including image classification, facial and object recognition, medical imaging analysis, and many more. In addition, there are applications such as physics-informed simulators in which accurate forecasts in real time with a minimal lag are required. The present neural network designs include millions of parameters, which makes it difficult to install such complex models on devices that have limited memory. Compression techniques might be able to resolve these issues by decreasing the size of CNN models that are created by reducing the number of parameters that contribute to the complexity of the models. We propose a compressed tensor format of convolutional layer, a priori, before the training of the neural network. 3-way kernels or 2-way kernels in convolutional layers are replaced by one-way fiters. The overfitting phenomena will be reduced also.  The time needed to make predictions or time required for training  using the original Convolutional Neural Networks model would be cut significantly if there were fewer parameters to deal with.  In this paper\footnote{\emph{This work has been submitted to Elsevier for possible publication.}} we present a method of  a priori compressing convolutional neural networks for finite element (FE) predictions of physical data. Afterwards we validate our a priori compressed models on physical data from a FE model solving a 2D wave equation. We show that the proposed convolutinal compression technique achieves equivalent performance as classical convolutional layers with fewer trainable parameters and lower memory footprint.
\end{abstract}

\clearpage

\section{Introduction}
Convolutional neural networks (CNNs) have become an effective tool for a wide range of tasks, such as image classification, object and facial recognition, medical image analysis, and many other things.
These applications are a good fit for them because of their capacity to recognize and learn from complicated patterns in data.
Although real-time forecasts with no lag are necessary for numerous applications such as autonomous driving cars and physics-informed simulators, the high number of parameters in existing CNN designs can make it challenging to deploy these models on systems with little memory.
In order to overcome this issue, compression algorithms have been developed to reduce the complexity and amount of parameters in CNN models, hence shrinking their size. In \cite{Gabor2022}  a tensor decomposition is propose to approximate N-way kernels involved in convolutional layers. But this compression step is applied a posteriori, after the training of the related neural network.

In this paper, we present an a priori method of compressing CNNs for finite element (FE) method physical data. This method relies on a neural network architecture that perform the decomposition of kernels in usual convolutional layer. This neural network archtecture involves an adequate processing of the data. Additionally, we explore approaches to optimize the data from FE models for CNN training. The efficiency of the compressed models is validated on physical data from a FE model solving a 2D wave equation. Our goal is to improve the efficiency of CNNs in real-world applications where computational resources are limited.

In many fields, the use of FE models is increasingly prevalent, as they provide a means of predicting the behavior of physical systems. For example, in the field of structural engineering, FE models are used to predict the response of structures to various loads, such as wind and earthquakes. By using CNNs to analyze FE data, we can gain insight into the behavior of these physical systems and make predictions that can guide engineering design and decision-making.

Compression techniques can be broadly divided into two categories: weight  pruning and weight quantization. Such compression are preformed a posteriori, after the training of the CNNs. Weight Pruning involves removing the least important weights or filters in a CNN, thereby reducing the number of parameters and computational complexity of the model. For example, in \cite{han2015learning}, the authors use pruning techniques to remove unimportant weights in a CNN, reducing the number of parameters with minimal loss of accuracy. In \cite{li2016pruning}, the authors propose a filter pruning method to remove unimportant filters from a pre-trained CNN, reducing the computational cost of the model without sacrificing accuracy.

Weight Quantization, on the other hand, involves reducing the precision of the weights in a CNN. This can be achieved by using fewer bits to represent each weight, thereby reducing the size of the model. In \cite{courbariaux2015binaryconnect}, the authors propose a method to quantize the weights of a CNN to binary values, reducing the number of parameters and computational cost of the model while maintaining accuracy. In \cite{rastegari2016xnor}, the authors propose a method to quantize the weights of a CNN to binary and ternary values, achieving a reduction in the number of parameters and computational cost of the model with minimal loss of accuracy.

In addition to these methods, there are other compression techniques, such as decomposition of weight matrices, that have been proposed to reduce the size of CNNs. In \cite{hameed2022convolutional}, the authors use Kronecker product decomposition to compress the weights of a CNN, reducing the number of parameters and computational cost of the model. A Tucker decomposition of 4-way kernels is proposed in \cite{Gabor2022}

Compression techniques can provide significant benefits in terms of reducing the size and computational cost of CNNs, making them more feasible for deployment on devices with limited memory and computational resources. Additionally, by reducing the number of parameters in a CNN, the risk of overfitting can be reduced, leading to improved generalization performance.

In this paper we propose a different approach by reversing the paradigm: instead of computing a decomposition of an already trained CNN, we propose to a priori construct the CNN as a succession of decomposed convolutional layers and learn each term of the decomposition by backpropagating gradients from the neural network training.

\section{Physical Data preprocessing}
\begin{figure}
\begin{center}
\includegraphics[height=5cm,keepaspectratio]{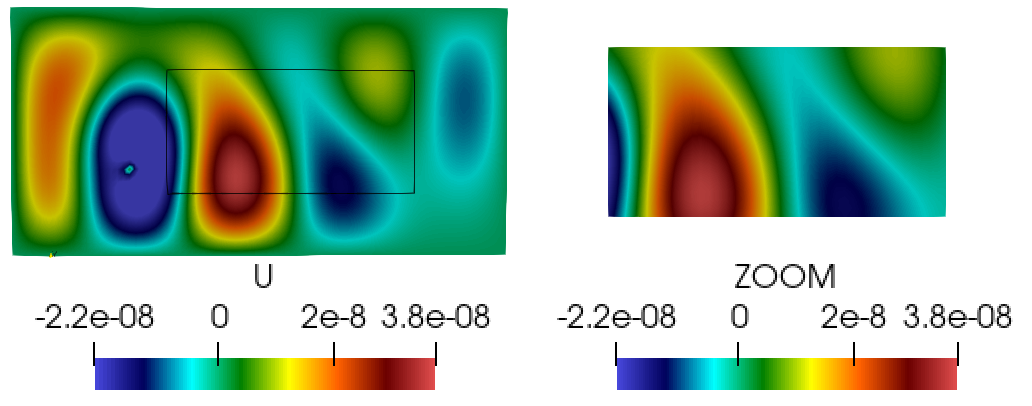}   
\caption{Visualization of the FE output on $\Omega$ and  $ \Omega ^{'}$} 
\label{fig:FEM}
\end{center}
\end{figure}

A boundary generator using a neural network is proposed, where physics-based equations are inforced in a submodel driven by a neural network. This neural network has to emulate the parametric predictions of a full finite element model, over a domain $\Omega$ in Figure \ref{fig:FEM} for instance. The submodel is a finite element model restrained to a zone of interest denoted by $\Omega^\prime$ in Figure \ref{fig:FEM}. It is supplemented by boundary conditions that are forecast by a neural network. This  method rely on the general principle which is a submodel formed of two weakly coupled components:
\begin{itemize}
\item A neural network learning boundary conditions around a  predetermined zone of interest, related to the submodel.
\item A finite element submodel in the zone of interest using boundary conditions generated by the neural network. We assume that there is no modeling error in the zone of interest covered by the proposed submodel.
\end{itemize}

For the validation of our approach, we choose to solve the 2D  wave equation. We define two 2D Cartesian space grids $\Omega_h$ and $\Omega'_h$, with $ \Omega'_h \subset \Omega_h$ representing the zone of interest.
$\Omega_h$ and $\Omega _h'$ are triangular space discretization of sizes $[N_x,N_y]$ and $[N'_x,N'_y]$ of the domains $[-L_x,L_x] \times  [-L_y,L_y]$  and  $[-L'_x,L'_x] \times  [-L'_y,L'_y]$. 
And finally a temporal grid $T$ is defined as discretization of size $N_T$ of the space $[0,T_{final}]$ and the time step $ \Delta t = \frac{T_{final}}{N_T-1}$ . The 2D wave equation is given as follows: 
\begin{equation} \left\{
\begin{array}{l}
\begin{aligned}
 & \frac{1}{c^2} \frac{\partial^2 u}{\partial t^2} - {\Delta u} = f ~~ on ~~  \Omega ~~ \forall t > 0  \\
&u = 0 ~~   on  ~~ \partial  \Omega ~~ \forall t > 0  \\
& u = u_{0} ~~   on  ~~  \Omega ~~ for  ~~ t  =  0
\end{aligned}
\end{array}
\right.
\end{equation}
where u is the amplitude of the wave. 
A source point is determined for the problem resolution where $(x_S,y_S)$ are the source point coordinates, it is chosen to be outside the zoom domain:
 i.e.  $(x_S,y_S) \in \Omega \backslash \Omega'$. \\
The source term at the right hand side of the wave equation is set as: 
\begin{equation} 
 (\forall t \in T),  ~~ f(x,y,t) =  sin(\omega t)  \delta_{(x_S, y_S)}(x,y) 
\end{equation}
where $\delta_{(x_S,y_S)}$ denotes the 2D Dirac distribution centered at $(x_S,y_S)$.
A three-dimensional parameter vector $  p = (\omega, x_s, y_s)$  is chosen and determined then sampled, (note that $c$ is constant over all samples since it is a parameter needed for the submodel). 
Sampling is done using latin hypercube sampling routines. For every parameter vector $p$ a simulation vector $U(p)$ is generated using the FE model.
One sample of data is then $ (p,U(p))$ where $ p \in D_p \subset  \mathbb{R}^3 $ and $ U(p)(t) \in V_h \subset \mathbb{R}^{N_X \times N_Y}.$  \\ 
Then, 2 datasets are generated from the same uniform distribution as the following:
\begin{itemize}
\item Training data set: 100 samples generated, used for training parametric approach models.
\item Test data set:  25 samples generated, used for testing the training process of each neural network, and comparing the models we used in our study.
\end{itemize}
All generated data are scaled with a standard scaler, and restrained to $\Omega^\prime$. These data are 3D (or 3-way) tensors that save the time-space evolution for each instance of input parameters in the FE model. There is 2 indices for 2D space coordinates, and one index for the time axis.

\section{Proposed Approach}
\subsection{Background}\label{cpdback}
Let us consider $W = (w_{i,o}) \in \mathbb{R}^{n_i,n_o}$ the weight matrix for a fully connected layer, bias vectors will be omitted in this section for simplification purpose, however, we can easily extend this approach by taking into account the bias. 
The output of a fully connected layer who takes as an input a vector $x \in \mathbb{R}^{n_i}$ is a vector $y \in \mathbb{R}^{n_o}$.
\begin{equation}
\left[ {\begin{array}{c}
    y_{1} \\
	\vdots \\
    y_{n_o} \\
  \end{array} } \right] = 
  \left[ {\begin{array}{ccc}
    w_{1,1} & \cdots & w_{1,n_i}\\
	\vdots  & \ddots & \vdots\\
    w_{n_o,1} & \cdots &  w_{n_o,n_i}\\
  \end{array} } \right] \times
\left[ {\begin{array}{c}
    x_{1} \\
	\vdots \\
    x_{n_i} \\
  \end{array} } \right]
\end{equation}

\noindent Where $W$ is a dense matrix, usually initialized as $ (\forall (i,o) \in [|1,n_i|] \times [|1,n_o|]) w_{i,o} \neq 0$.

In \cite{mallat2016understanding} 2D convolutional layers are defined as follows, given an input $x \in \mathbb{R}^{n_x \times m_x}$ the output of the convolution of $x$ with a kernel  $K \in \mathbb{R}^{k_1 \times k_2}$ is an 2D matrix $y \in  \mathbb{R}^{n_y \times m_y}$ :
\begin{equation}\label{eq:convfilt}
y(i,j) = (x \star K)(i,j) = \sum_{l=1}^{k_1}\sum_{h=1}^{k_2} x(l+i-1,h+j-1)K(l,h)
\end{equation}
This approach can be generalized for higher dimension inputs with N-ways kernels ($N > 2$). Thus, since a convolutional layer consists of a linear combination of the outputs of multiple convolution kernels passed in parameter for a non linear activation function $\sigma$, we can write for an input $x \in \mathbb{R}^{c \times n_x \times m_x}$, c being the numbers of channels in the input,  and a collection of kernels $K \in \mathbb{R}^{c_k \times n_k \times m_k}$ where $n_k \leq n$ and $m_k \leq m$ and $c_k$ the number of Kernels defines the number of channels for the output, which is usually called the feature map, which is a tensor $y  \in \mathbb{R}^{c_k \times n_y \times m_y}$ :
\begin{equation}
(\forall j \in [|1,c_k|]) ~~ y[j,:,:] = \sigma(\sum_{i=1}^{c} x[i,:,:] \star K[j,:,:])
\end{equation}

For visualization let us consider the example of one kernel ($c_k=1$), where $k_1 = k_1 = 2$, $n_i = n_o = 3$, developpements in this section are written in the case of 2D convolutional kernels but can easily be generalized N-way kernels ($N>2$). Having :
$K = \left[ {\begin{array}{cc}
    k_{1,1} & k_{1,2} \\
	k_{2,1} & k_{2,2} \\
  \end{array} } \right]$
and $ x = \left[ {\begin{array}{ccc}
    x_{1,1} & x_{1,2} & x_{1,3} \\
	x_{2,1} & x_{2,2} & x_{2,3}\\
	x_{3,1} & x_{3,2} & x_{3,3}\\
  \end{array} } \right]$, and considering the flattened vector of x, the flattened output of this convolutional layer is a vector $y \in \mathbb{R}^4$:

\begin{equation}\label{sparseconv}
\left[ {\begin{array}{c}
    y_{1,1} \\
	y_{1,2} \\
	y_{2,1} \\
	y_{2,2} \\
  \end{array} } \right] = 
  \left[ {\begin{array}{ccccccccc}
    k_{1,1} & k_{1,2} & 0 & k_{2,1} & k_{2,2} & 0 & 0 & 0 &0 \\
	0 & k_{1,1} & k_{1,2} & 0 & k_{2,1} & k_{2,2} & 0 & 0 & 0\\
	0 & 0 & 0 & k_{1,1} & k_{1,2} & 0 & k_{2,1} & k_{2,2} & 0\\
	0 & 0 & 0 & 0& k_{1,1} & k_{1,2} & 0 & k_{2,1} & k_{2,2}\\
  \end{array} } \right] \times
\left[ {\begin{array}{c}
    x_{1,1} \\
	x_{1,2} \\
    x_{1,3} \\
	x_{2,1} \\
	x_{2,2} \\
    x_{2,3} \\
	x_{3,1} \\
	x_{3,2} \\
    x_{3,3} \\
  \end{array} } \right]
\end{equation}
  
Thus convolutional layers can be considered as sparse fully connected layers that have a lower computational complexity than dense layers.

\subsection{Canonical Polyadic Decomposition of convolutional kernels}
Canonical Polyadic Decomposition (CPD) \cite{kolda2009tensor,9733397,hitchcpd} describes a N-way tensor as a sum of rank one tensors. In contrast to the matrix scenario, the CPD of a low rank tensor is unique given mild assumptions. CPD's intrinsic distinctiveness makes it a strong tool in many applications, allowing for the extraction of component information from a signal of interest. The generalized eigenvalue decomposition (GEVD), which picks a tensor matrix subpencil and then computes the generalized eigenvectors of the pencil, is a common approach for algebraic calculation of a CPD.

\begin{figure}[H]
\begin{center}
\includegraphics[width=12cm,keepaspectratio]{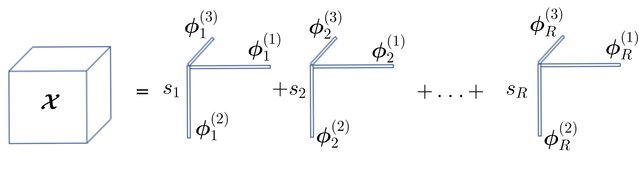}
\caption{Canonical Polyadic Decomposition from \cite{pham2018computational}} 
\label{fig:pb}
\end{center}
\end{figure}

Let us consider a CPD of the convolutional kernel k defined in Section \ref{cpdback} with 2 vectors $Q,R$:

\begin{equation}\label{cpdfilter}
K \approx Q \otimes R^{T} = \left[ {\begin{array}{c}
    q_{1,1} \\ 
	q_{1,2} \\
  \end{array} } \right] \otimes \left[ {\begin{array}{cc}
    r_{1,1} & r_{2,1} \\
  \end{array} } \right] =  \left[ {\begin{array}{cc}
    q_{1,1} r_{1,1} & q_{1,1} r_{2,1} \\
	q_{1,2} r_{1,1} & q_{1,2} r_{2,1} \\
  \end{array} } \right]
\end{equation}

\noindent Where $\otimes$ denotes the outer vector product. Determining the values of $Q,R$ will be discussed in a following section. Thus, considering Equation \eqref{cpdfilter},  the Equation \eqref{sparseconv} can be rewritten:

\begin{equation}
\begin{aligned}
 & \left[ {\begin{array}{c}
    y_{1,1} \\
	y_{1,2} \\
	y_{2,1} \\
	y_{2,2} \\
  \end{array} } \right] = \mathcal{Q} \times \mathcal{R} \times x =
  \left[ {\begin{array}{cccccc}
    q_{1,1}  & 0  & q_{1,2} & 0 & 0  & 0  \\
	0 & q_{1,1} & 0  & q_{1,2} & 0  &0\\
	0  & 0  & q_{1,1} & 0 & q_{1,2} &0  \\
	0 & 0 & 0  & q_{1,1} & 0 & q_{1,2}  \\
  \end{array} } \right] \times \\
&  \left[ {\begin{array}{ccccccccc}
    r_{1,1}  & r_{2,1}  & 0 & 0 & 0  & 0  & 0 & 0 & 0\\
	0 & r_{1,1} & r_{2,1}  & 0 & 0 & 0  & 0  & 0 & 0\\
	0 & 0 & 0 & r_{1,1} & r_{2,1}& 0 & 0  & 0  & 0\\
	0 & 0 & 0 & 0 & r_{1,1} & r_{2,1}& 0 & 0  & 0  \\
	0 & 0 & 0  & 0 & 0 & 0 & r_{1,1}  & r_{2,1}  & 0\\
	0 & 0 & 0 & 0  & 0 & 0 & 0 & r_{1,1}  & r_{2,1} \\
  \end{array} } \right] \times
 \left[ {\begin{array}{c}
    x_{1,1} \\
	x_{1,2} \\
    x_{1,3} \\
	x_{2,1} \\
	x_{2,2} \\
    x_{2,3} \\
	x_{3,1} \\
	x_{3,2} \\
    x_{3,3} \\
  \end{array} } \right]
\end{aligned}
\end{equation}

Where $\mathcal{R}$ is the matrix of the linear operation of applying the kernel $R$ to every row of $x \in \mathbb{R}^{3 \times 3}$ and $\mathcal{Q}$ is the matrix of the linear operation of applying the kernel $Q$ to every column of the output of applying $R$ to $x$, then we have rewritten the 2D convolutional layer of $x$ by the kernel $K$ to successfully applying two 1D kernels, induced by the CPD of K, to different dimensions of $x$.

\subsection{Approximation of the CPD of convolutional layers}
In the previous section we presented an approach to reduce the dimensionality of a trained convolutional layer using CPD of each kernel, the main issue remaining is determining the components of each decomposition. Generalized Eigenvalue Decomposition  \cite{Domanoveigen} is the default go-to approach. In this work we propose an equivalent approach relying on Singular Values Decomposition of kernels. 
Let us first consider the case of 2D kernels, let $K \in \mathbb{R}^{n_1 \times n_2}$ be a 2D kernel. The SVD  of K is written as :
\begin{equation}\label{eq:svdk}
K \approx \sum_{i=1}^r \sigma_i K^{(L)}_i \otimes K^{(R)}_i.
\end{equation}

Where $r$ is the number of singular values considered, $(\sigma_i)_{(i \leq r)}$ the singular values and $(K^{(L)}_i, K^{(R)}_i)$ respectively the left and right singular vectors.
Equation \eqref{eq:svdk} can be rewritten as:

\begin{equation}\label{eq:svdk2d}
K \approx \sum_{i=1}^r K'^{(L)}_i \otimes K'^{(R)}_i.
\end{equation}
Where each vector $K'^{(.)}_i = \sqrt{\sigma_i} K^{(.)}_i$, and by construction, $(K'^{(L)}_i, K'^{(R)}_i)$ are rank one vectors, thus we obtained a Polyadic Decomposition of a 2D kernel. Depending on $r$, this decomposition may fulfill a precision criteria which is sufficient for this application.

Let us now consider the case of 3D kernels, let $K \in \mathbb{R}^{n_1 \times n_2 \times n_3}$ a 3-way tensor, and let us consider $K_l = K[:,:,l]$, K can be written as:
\begin{equation}
K = \sum_{l=1}^{n_3} K_l \otimes I_l
\end{equation} 
Where $(I_l)_{l \leq n_3}$ are the vector rows of the identity matrix in $\mathbb{R}^{n_3 \times n_3}$, and each $(K_l)_{l \leq n_3}$ is a 2D matrix, thus a decomposition in the form of \eqref{eq:svdk2d} is possible.
Therefore K can be decomposed as:
\begin{equation}
K = \sum_{l=1}^{n_3} K_l \otimes I_l \approx \sum_{l=1}^{n_3}\sum_{i=1}^r K'^{(L)}_{i,l} \otimes K'^{(R)}_{i,l} \otimes I_l.
\end{equation}
With straightforward work on sums indexes, one can rewrite this approximation as a single  sum of outer products of rank one vectors. Thus, we define a recursive method to build decompositions of a tensor of any dimension.

\subsection{A priori decomposed convolutional layers}
Applying this decomposition formalism to every kernel of convolutional layer leads to a compressed representation of the convolutional layer. We show that the a priori decomposition requires intermediate reshape and transpose operations on data.  The new convolution approach is detailled in what follows.

Let us first consider $X \in \mathbb{R}^{n_B \times n_c \times n_t \times n_x}$ a tensor of physical 2D data, formed by $n_B$ instances (or samples of data) and $n_c$ channels (or features). In our target application, for each sample, $n_t$ is the size of the temporal dimension of the data and $n_x$ the size of the spatial dimension. Let $K \in \mathbb{R}^{n_f \times n_{kt} \times n_{kx}}$ where $n_f$ is the number of 2D kernels considered, $n_{kt}$ and $n_{kx}$ respectively the kernels sizes in temporal and spatial dimension, and let $Y,\beta \in \mathbb{R}^{n_B \times n_f \times n'_t \times n'_x}$ being respectively the output and the biases of the convolutional layer. Using the equation \eqref{eq:convfilt} defining the output of a convolution by a kernel we define the output $Y$ of the convolution of $X$ by $K$ and $\beta$ as :
\begin{equation}
Y(i,j,h,l) =  \sum_{v=1}^{n_f} \sum_{o = 1}^{n_{kt}} \sum_{p = 1}^{n_{kx}} X(i,v,h+o-1,l+p-1)K(j,o,p) + \beta(i,j,h,l)
\end{equation}
 
In these developpements we consider the case where all inputs channels are convolved with all output channels and default values are set for other convolution parameters $(stride = 1, ~padding = 0,~ dilation = 1)$,  generalizing these developpements for generic values or for higher dimension convolutional layers is a straightforward work on indexes or data structure. Let us consider a formal SVD of each kernel in K and only focusing on the first term of the sum (using more singular values for the decomposition can be performed by using more channels in the decomposed kernels):
\begin{equation}
j=1, ~~ K[j,:,:] =  K^t_j \otimes K^x_j + R[j,:,:]
\end{equation} 
where $R$ is the residual tensor of the rank one approximation of $K$.
Thus we can extract two subsets of 1D kernels forming 2 differents temporal and spatial 1D convolutional layers $K_j^t \in \mathbb{R}^{n_t}$ and $K_j^x \in \mathbb{R}^{n_x}$.
Let us consider $\tilde{Y} \in \mathbb{R}^{n_B \times n_c \times n_t \times n'_x}$ the output of the spatial convolutional layer $K_j^x$ on the columns of $X$ without combining the ouptuts, we obtain:
\begin{equation}
\tilde{Y}(i,j,h,l) = \sum_{p = 1}^{n_{kx}} X(i,(j,h),l+p-1)K_j^x(p)
\end{equation}
where $(j,h)$ is a multi index obtained by reshaping the data. In the sequel, the following transpose operation is also required:
\begin{equation}
\tilde{Y}^T(i,j,l,h) = \tilde{Y}(i,j,h,l)
\end{equation}
Let us now consider $\hat{Y} \in \mathbb{R}^{n_B \times n_f \times n'_t \times n'_x}$ the output of the temporal convolutional layer $K_j^t$ on the lines of $\tilde{Y}$ and combining the ouptuts and then adding the same biases $\beta \in \mathbb{R}^{n_B \times n_f \times n'_t \times n'_x}$, we obtain:

\begin{equation}
\begin{aligned}
\hat{Y}(i,\hat{j},h,l) & =  \sum_{j=1}^{n_f} \sum_{o = 1}^{n_{kt}} \tilde{Y^T}(i,(j,l),h+o-1)K_j^t(o) + 						\beta(i,\hat{j},h,l) \\
				 & = \sum_{j=1}^{n_f} \sum_{o = 1}^{n_{kt}}\left( \sum_{p = 1}^{n_{kx}} X(i,j,h+o-1,l+p-1)K_j^x(p)\right) K_j^t(o) + \beta(i,\hat{j},h,l) \\
				 & = \sum_{j=1}^{n_f} \sum_{o = 1}^{n_{kt}} \sum_{p = 1}^{n_{kx}} X(i,j,h+o-1,l+p-1) \: \left( K_j^x(p)\: K_j^t(o)\right) + \beta(i,\hat{j},h,l) 
\end{aligned}
\end{equation}
where $K_j^x(p)\: K_j^t(o)$ is the rank-one tensor approximation of the 2D convolution kernel. It follows that:
\begin{equation}
\begin{aligned}
Y(i,\hat{j},h,l) & - \hat{Y}(i,\hat{j},h,l)  = \\ 
& \sum_{j=1}^{n_f} \sum_{o = 1}^{n_{kt}} \sum_{p = 1}^{n_{kx}} X(i,j,h+o-1,l+p-1) \: R(j,o,p) 
\end{aligned}
\end{equation}
Therefore the smaller $R$, the smaller the discrepancy between $Y$ and $\hat{Y}$.

Thus, using  decompositions of kernels of a 2D convolutinal layers we obtained an approximation of the output of said layer by using the decomposed kernels in two consecutive 1D convolutinal layers with index manipulations via data reshaping and transpose operations. 

Similarly, the rank-one decomposition can be extended to d-way kernels:
\begin{equation}
\begin{aligned}
\hat{Y}(i,\hat{j},h_1,\ldots,h_d)  = & \sum_{j=1}^{n_f} \sum_{p_1 = 1}^{n_{k1}} \ldots \sum_{p_d = 1}^{n_{kd}} \\
& \quad X(i,j,h_1+p_1-1,\ldots, h_d+p_d-1)\\
&\quad  \Pi_{k=1}^d K_j^k(p_k) \\
& + \beta(i,\hat{j},h_1,\ldots,h_d)
\end{aligned}
\end{equation}

Thus reducing the number of trainable parameters of the convolutional layer from $\Pi_{k=1^d n_{kk}}$  to $\sum_{k=1}^d n_{kk}$. Therefore we decomposed a model which complexity is exponentialy dependent on the dimension of the problem to a model which complexity is linealy dependent on the dimension, thus alienating the curse of dimensionality.
In addition, since we approximate d-way kernels with a decomposition of d 1D convolutional layer, we can consider each layer appart and use activation functions after each 1D convolution, thus constructing a non linear decomposition of d-way kernels.
Indeed this a priori decomposition of CNNs assume that parameters on each kernel are dimensionaly separable, nevertheless the update of those parameters while training the CNN depends on every dimension of the problem solved since the gradients backpropagated depends on all dimensions, then the dimensional separability of the problem is not a necessary hypothesis to perform such decomposition. Nevertheless, the precision of such decomposition is not an indicator of the precision of training models using this new form of convolutions. But it will rather be inforced by solving the optimization problem related to the objective function.

\section{Weight sharing}

Weight sharing or layer coupling is a deep learning model order reduction method in which multiple models which objective is to extract different features from same inputs share the first extraction layers. In \cite{xie2021weight} the authors present an overview of different implementation and optimization method of weight sharing, as for \cite{pmlrv80pham18a} in which the authors present an approach to builds a large computational graph with each subgraph representing a neural network design, requiring all architectures to share their parameters. A policy gradient is used to train a controller to find the subgraph that maximizes the reward on a validation set. \cite{liu2016coupled} proposes a weight sharing approach for Generative Adversarial Network to learn joint distribution.

In our approach, weight sharing is a relevent choice, since the objective is to predict different physical fields with respect to the same input parameters. Thus, instead of training two different models, we train coupled models, therefore reducing the number of parameters in the first layers by half. The weight of the shared layers are updated with gradients induced by the minimization of the empirical risk for all physical fields predicted, the last regressive layers are updated by gradients induced by the only physical field predicted. 

Let us denote $M_d,~M_v$ the models trained to approach the displacement and its time derivative (the velocity) fields using weight sharing, let us denote $N_s$ the operation of applying the shared layers on a parameters vector input $p$, and $N_d,~N_v$  the operation of applying the remaining layers for the displacement and velocity models to the outputs of $N_s$. Outputs of both models can be written as:
\begin{equation}
\left\{
\begin{array}{c}
M_d(p) = N_d(N_s(p)) \\ 
M_v(p) = N_v(N_s(p))
\end{array}
\right.
\end{equation}

\section{Time regularization}
In our approach, since the objective is to predict dynamic physical fields, predicting fields that are dynamically linked is quite frequent, as for example predicting displacement and velocity fields. We propose an approach for time regularization of the predicted fields by adding a residual minimization in the cost function, which is quite similar to the approaches in Physics Informed Neural Network \cite{PINN} in which the residual of the Partial Differential Equation is minimized. Since for FE models access to the residual of the PDE solved is intrusive and often infeasible, we rely on the temporal regularity of the approximated physical fields.

Let us consider $P$ and $T$ sets of collocation points for the time residual computing, and let $M_u$ and $M_v$ be two neural networks which objective is to predict displacement and velocity fields for the same FE models parameterized by $p$. We define the time residual as :
\begin{equation}
\E\limits_{p \in P} \E\limits_{t' \in T}\lvert| \frac{\partial M_u(p,t')}{\partial t} - M_v(p,t')\rvert|_2^2
\end{equation} 
For convolutional layers, computing the derivative of the output with respect to the inputs is not always feasible, so we approach the time derivative with an Euler finite difference scheme. In the following this regularization will be mentionned as the Euler regularization.

\section{Developped models}
All models in this following work can be categorized into 9 categories depending on how they process each type of physical data and how spatial and temporal information is used to update the weights at each training step. So for spatial and temporal information we distinguish 3 categories each:
\begin{itemize}
\item Sampled: only one sample of spatial or/and temporal data is used to update the weights, the models takes as inputs the spatial or temporal coordinates or both.
\item Local: only a local amount of spatial or/and temporal data is used to update the weights, the model predicts the whole simulation but uses convolutional layers to treat data localy.
\item Global: the whole information across the spatial or temporal information or both is used to update the weights, this is achieved in a convolutional layer by considering the global dimension as the channels in the case where the other dimension is not globaly processed, or with a fully connected layer.
\end{itemize}

\begin{figure}[H]
\begin{center}
\includegraphics[height=8cm,keepaspectratio]{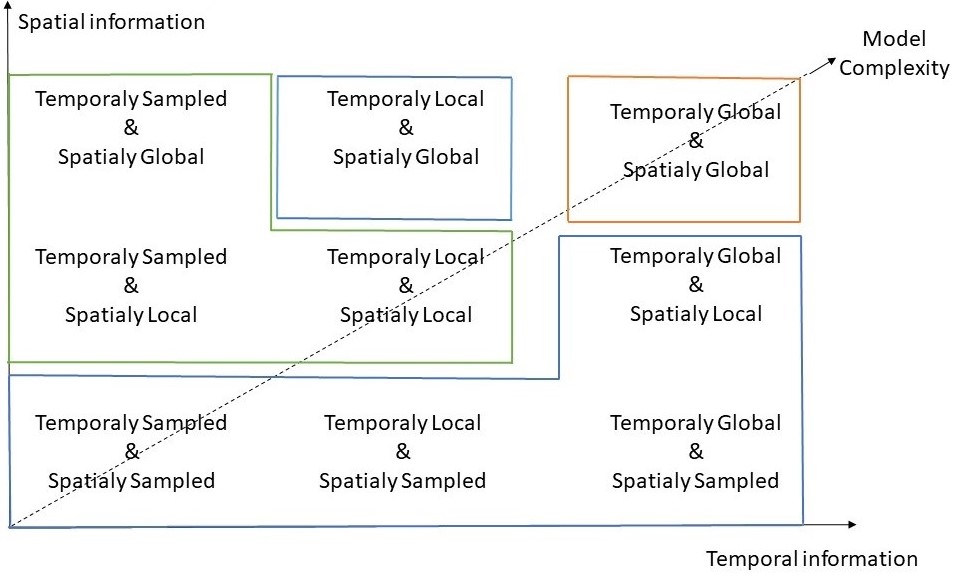}
\caption{Model complexity evolution} 
\label{fig:models}
\end{center}
\end{figure}

Figure \ref{fig:models} shows the evolution of model complexity according to the approaches considered spatialy and temporally, in green approaches that were considered and developed in this work, in blue feasible approaches but not adapted to our application and in orange non-feasible approaches.

\subsection{Models annotations}
Regarding our developed model, time sampled approaches or time conditioned approaches will be suffixed by $"\_t"$, spatialy local approaches are the convolutional networks that will be prefixed by $"Conv"$ followed by their dimension, local and global temporal approaches can be distinguished by the dimension of the convolutional net, for example using a 3D convolutional network to approach a 2D spatial and 1D physical field is temporally local, using a 2D convolutional layer to approach the same field is temporally global. All fully connected models developed are spatialy global and temporally sampled. All of these models can have multiple variants considering if convolutional decomposition is performed a suffixe "N.5D" means that a convolutional layer of dimension $N+1$ has been decomposed using the a priori decomposition approach presented in this paper, $2.5D$ means that the model has been decomposed to a 2D spatial layer and a 1D temporal layer, $2.5Db$ means that a 3D model has been a priori decomposed to three 1D layers. Additional variants appears depending on the regularization technique used, Batch Normalization \cite{batchnorm} is designated by $"BN"$, Euler regularization by $"E"$, Weight sharing or layer sharing by $"SL"$, $"BASIC"$ will denote a network with no regularization applied. All these type of regularization can be combined except for Batch Normalization and Euler, since Batch Normalization makes the derivatives computed in Euler Regularization erroneous.

\section{Numerical results}
\subsection{Error indicator}
We define a relative error indicator over the time and space grid of the interest zone, in order to quantify the precision of our submodels as $\epsilon$.
For a submodel M, a parameter vector p, and a time value t:
\begin{equation}
\epsilon (M,p,t) =\frac{\E\limits_{(x,y) \in \Omega '} [ \displaystyle\left\lvert M(p)(t,x,y)  - U(p)(t,x,y) \right\rvert ]}{\max\limits_{x,y \in \Omega '} \displaystyle\left\lvert U(p)(t,x,y) \right\rvert } 
\end{equation}

For a comparison over the testing data set: 
\begin{equation}
\epsilon (M,t) =\E\limits_{p \in \mathbb{P}_{Test}} [\epsilon (M,p,t)] 
\end{equation}
\subsection{2D Wave propagation with one source point and early stopping}
We train all regression models defined in the previous section on FE prediction in $\Omega^\prime$ as training data, for 1000 epoch using Adam \cite{Adam} optimizer and a learning rate of $1e^{-3}$. In all tables, results in bold black indicate best results in terms of accuracy, results in blue indicate results within an acceptable threshold, and results in red indicate unacceptable results.
\input{Tables/trainp3.tex}

Table \ref{trainp3} shows the number of trainable parameters for each model combined with every regularization technique described before. As expected fully connected layers and approaches using temporal dimension as feature maps have the highest number of parameters, and our decomposition approach is efficient in reducing the number of trainable parameters in each model. Sharing Layers approach reduce drastically the number of parameters in convolutional layer, fully connected layers cannot benefit from this approach since first layers have the lowest number of parameters. Batch norm regularization only add a negligible number parameters.
\input{Tables/trainuf2.tex}
\input{Tables/trainvf2.tex}
\input{Tables/testuf2.tex}
\input{Tables/testvf2.tex}

Tables \ref{TrainUF2} and \ref{TrainVF2} show respectively the training error of each model variant for data prediction on $\Omega^\prime$ considering all the possible regularization methods described in the previous sections. The tables indicate that the model that achieves the best accuracy in training is $Conv3D$ with BatchNorm regularization and achieves good generalization error as shown by  Tables \ref{TestUF2} and \ref{TestVF2} where best compromise between training error and generalization error is achieved by $Conv3D$, this behavior can be explained by the overfitting occurring in $Conv2D$ considering the large amount of parameters within the model. Tables \ref{TrainUF2}, \ref{TrainVF2}, \ref{TestUF2} and \ref{TestVF2} show that the approaches using our convolutional decomposition presented in this paper achieve comparable training precision and generalization error with much fewer parameters as shown by Table \ref{trainp3}. Only the models with the Batch Normalization regularization could achieve acceptable error threshold for training with few epoch and a high learning rate. Euler regularization shows no improvements on the precision of the models, however combined with sharing layers it achieves equivalent error threshold to the Batch Normalization results.

\input{Tables/trainufz2.tex}
\input{Tables/trainvfz2.tex}
\input{Tables/testufz2.tex}
\input{Tables/testvfz2.tex}
Tables \ref{TrainUFZ2}, \ref{TrainVFZ2}, \ref{TestUFZ2} and \ref{TestVFZ2} show respectively the training error and generalization error of each model variant after the submodeling operation in the area of interest considering all the possible regularization methods described in the previous sections. The tables indicate that the model that achieves the best training and generalization are boundary generative models, achieving this results with even fewer trainable parameters, the convolutional decomposition approaches achieve equivalent results with fewer trainable parameters.

\input{Tables/trainepoch.tex}

The table \ref{trainepoch} shows the average time for each model to train for one epoch, it indicates that the convolutional decomposition approaches have higher training time, this can be explained by comparing our first implementation with optimized implementation of deep learning libraries, the objective to attain comparable results with decomposed convolutional layers was achieved but the optimization of training time is yet to be achieved.

\subsection{2D wave propagation with one source point}
We train all models defined in the previous section for 10000 epoch using Adam \cite{Adam} optimizer and a learning rate of $1e^{-3}$ and learning rate decay to achieve a learning rate of $1e^{-4}$ at the last epoch.
\input{Tables/trainuf.tex}
\input{Tables/trainvf.tex}
\input{Tables/testuf.tex}
\input{Tables/testvf.tex}

Tables \ref{TrainUF} and \ref{TrainVF} shows respectively the training error of each model variant for data prediction considering all the possible regularization methods described in the previous sections. The tables indicate that the model that achieves the best accuracy in training is $Conv2D$ with BatchNorm regularization, but achieves poor generalization error as shown by  Tables \ref{TestUF} and \ref{TestVF} where best compromise between training error and generalization error is achieved by $Conv3D$, this behavior can be explained by the overfitting occuring in $Conv2D$ considering the large amount of parameters within the model. Tables \ref{TrainUF}, \ref{TrainVF}, \ref{TestUF} and \ref{TestVF} show that the approaches using the convolutional decomposition presented in this paper achieve comparable training precision and generalization error with much fewer parameters as shown by Table \ref{trainp3}. Results here show that all model could converge thanks to the higher number of epochs and learning rate decay, thus combining Batch Normalization within our decomposition approach allow the models to learn and have good generalization error with high learning rate and few epochs.
 
\input{Tables/trainufz.tex}
\input{Tables/trainvfz.tex}
\input{Tables/testufz.tex}
\input{Tables/testvfz.tex}
Tables \ref{TrainUFZ}, \ref{TrainVFZ}, \ref{TestUFZ} and \ref{TestVFZ} show respectively the training error and generalization error of each model variant after the submodeling operation in the area of interest considering all the possible regularization methods described in the previous sections. The tables indicate equivalent results  from the previous experiment, in the exception of all models achieve equivalent performance as previously explained. Models relying on Euler regularization achieve the best error threshold.

\section{Conclusion}
In this work, we presented  a novel method of compressing convolutional neural networks for FE physical data and approaches to optimize data from FE models for CNN training. Our compression approach can also be applied to learning data in higher dimensions since the complexity of the models is linearly dependent on the dimension and actual deep learning code library only allow up to 3D data learning. After that, we validated our compressed models on physical data derived from a FE model that was used to solve a 2D wave equation, we combined each approach with different regularization approaches, and showed that our convolutinal compression technique achieves equivalent performance as classical convolutional layers with fewer trainable parameters, 7 millions parameters for the classical approach versus 1 million parameters for  the decomposed approach.

\bibliographystyle{apalike}
\bibliography{bibcnn.bib}

\end{document}

%% file: Tables/trainp3.tex
\begin{table}[H]
\caption{Number of trainable parameters (million parameters)}\begin{center}
\begin{tabular}{|c|c|c|c|c|c|c|}
\hline 
& Basic & BN & E & SL & BN \& SL & E \& SL \\ \hline
FC t & \textcolor{red}{14.34} & \textcolor{red}{14.37} & \textcolor{red}{14.34} & \textcolor{red}{14.22} & \textcolor{red}{14.26} & \textcolor{red}{14.22} \\ \hline
Conv2D & \textcolor{red}{23.30} & \textcolor{red}{23.31} & \textcolor{red}{23.30} & \textcolor{red}{12.63} & \textcolor{red}{12.64} & \textcolor{red}{12.63} \\ \hline
Conv2D t & \textcolor{red}{1.52} & \textcolor{red}{1.52} & \textcolor{red}{1.52} & \textcolor{red}{0.81} & \textcolor{red}{0.81} & \textcolor{red}{0.81} \\ \hline
Conv3D & \textcolor{red}{7.25} & \textcolor{red}{7.25} & \textcolor{red}{7.25} & \textcolor{red}{3.83} & \textcolor{red}{3.83} & \textcolor{red}{3.83} \\ \hline
Conv2.5D & \textcolor{red}{2.72} & \textcolor{red}{2.72} & \textcolor{red}{2.72} & \textcolor{red}{1.46} & \textcolor{red}{1.46} & \textcolor{red}{1.46} \\ \hline
Conv2.5Db & \textcolor{red}{1.63} & \textcolor{red}{1.64} & \textcolor{red}{1.63} & \textcolor{red}{0.85} & \textcolor{red}{0.86} & \textcolor{red}{0.85} \\ \hline
FC t Boundary & \textcolor{blue}{0.27} & \textcolor{blue}{0.28} & \textcolor{blue}{0.27} & \textcolor{blue}{0.26} & \textcolor{blue}{0.26} & \textcolor{blue}{0.26} \\ \hline
Conv1D Boundary & \textcolor{red}{6.98} & \textcolor{red}{6.99} & \textcolor{red}{6.98} & \textcolor{red}{3.65} & \textcolor{red}{3.66} & \textcolor{red}{3.65} \\ \hline
Conv1D t Boundary & \textcolor{blue}{0.45} & \textcolor{blue}{0.45} & \textcolor{blue}{0.45} & \textbf{0.24} & \textcolor{blue}{0.24} & \textbf{0.24} \\ \hline
Conv2D Boundary & \textcolor{red}{2.19} & \textcolor{red}{2.19} & \textcolor{red}{2.19} & \textcolor{red}{1.13} & \textcolor{red}{1.13} & \textcolor{red}{1.13} \\ \hline
Conv1.5D Boundary & \textcolor{red}{1.18} & \textcolor{red}{1.19} & \textcolor{red}{1.18} & \textcolor{red}{0.61} & \textcolor{red}{0.62} & \textcolor{red}{0.61} \\ \hline
\end{tabular}
\label{trainp3}
\end{center}
\end{table}

%% file: Tables/trainuf2.tex
\begin{table}[H]
\caption{Train error on full displacement}\begin{center}
\begin{tabular}{|c|c|c|c|c|c|c|}
\hline 
& Basic & BN & E & SL & BN \& SL & E \& SL \\ \hline
FC t & \textcolor{red}{2.27} & \textcolor{red}{0.54} & \textcolor{red}{2.28} & \textcolor{red}{2.02} & \textcolor{red}{0.55} & \textcolor{red}{2.02} \\ \hline
Conv2D & \textcolor{red}{4.61} & \textbf{0.22} & \textcolor{red}{4.86} & \textcolor{red}{4.90} & \textcolor{blue}{0.43} & \textcolor{red}{4.66} \\ \hline
Conv2D t & \textcolor{red}{26.91} & \textcolor{blue}{0.41} & \textcolor{red}{26.91} & \textcolor{red}{26.91} & \textcolor{blue}{0.31} & \textcolor{red}{26.92} \\ \hline
Conv3D & \textcolor{red}{5.87} & \textcolor{blue}{0.24} & \textcolor{red}{5.76} & \textcolor{red}{6.08} & \textcolor{blue}{0.27} & \textcolor{red}{5.86} \\ \hline
Conv2.5D & \textcolor{red}{4.64} & \textcolor{blue}{0.34} & \textcolor{red}{5.11} & \textcolor{red}{4.55} & \textcolor{blue}{0.37} & \textcolor{red}{5.35} \\ \hline
Conv2.5Db & \textcolor{red}{4.66} & \textcolor{red}{0.54} & \textcolor{red}{5.37} & \textcolor{red}{4.27} & \textcolor{red}{0.79} & \textcolor{red}{0.61} \\ \hline
\end{tabular}
\label{TrainUF2}
\end{center}
\end{table}

%% file: Tables/trainvf2.tex
\begin{table}[H]
\caption{Train error on full velocity}\begin{center}
\begin{tabular}{|c|c|c|c|c|c|c|}
\hline 
& Basic & BN & E & SL & BN \& SL & E \& SL \\ \hline
FC t & \textcolor{red}{2.26} & \textcolor{red}{0.63} & \textcolor{red}{2.42} & \textcolor{red}{2.27} & \textcolor{red}{0.60} & \textcolor{red}{2.25} \\ \hline
Conv2D & \textcolor{red}{4.30} & \textbf{0.18} & \textcolor{red}{4.55} & \textcolor{red}{4.55} & \textcolor{blue}{0.22} & \textcolor{red}{4.31} \\ \hline
Conv2D t & \textcolor{red}{21.69} & \textcolor{blue}{0.25} & \textcolor{red}{21.69} & \textcolor{red}{21.68} & \textcolor{blue}{0.27} & \textcolor{red}{21.69} \\ \hline
Conv3D & \textcolor{red}{6.06} & \textcolor{blue}{0.23} & \textcolor{red}{5.83} & \textcolor{red}{6.02} & \textcolor{blue}{0.26} & \textcolor{red}{5.96} \\ \hline
Conv2.5D & \textcolor{red}{4.44} & \textcolor{blue}{0.25} & \textcolor{red}{4.78} & \textcolor{red}{4.33} & \textcolor{red}{0.43} & \textcolor{red}{4.81} \\ \hline
Conv2.5Db & \textcolor{red}{4.47} & \textcolor{red}{0.63} & \textcolor{red}{4.80} & \textcolor{red}{4.18} & \textcolor{red}{0.58} & \textcolor{red}{0.54} \\ \hline
\end{tabular}
\label{TrainVF2}
\end{center}
\end{table}

%% file: Tables/testuf2.tex
\begin{table}[H]
\caption{Test error on full displacement}\begin{center}
\begin{tabular}{|c|c|c|c|c|c|c|}
\hline 
& Basic & BN & E & SL & BN \& SL & E \& SL \\ \hline
FC t & \textcolor{red}{1.34} & \textcolor{red}{0.37} & \textcolor{red}{1.37} & \textcolor{red}{1.24} & \textcolor{red}{0.46} & \textcolor{red}{1.29} \\ \hline
Conv2D & \textcolor{red}{3.25} & \textcolor{blue}{0.24} & \textcolor{red}{3.15} & \textcolor{red}{3.22} & \textcolor{blue}{0.31} & \textcolor{red}{3.46} \\ \hline
Conv2D t & \textcolor{red}{24.45} & \textcolor{blue}{0.28} & \textcolor{red}{24.46} & \textcolor{red}{24.47} & \textcolor{blue}{0.29} & \textcolor{red}{24.47} \\ \hline
Conv3D & \textcolor{red}{4.72} & \textbf{0.18} & \textcolor{red}{4.66} & \textcolor{red}{4.64} & \textcolor{blue}{0.26} & \textcolor{red}{4.81} \\ \hline
Conv2.5D & \textcolor{red}{3.47} & \textcolor{blue}{0.28} & \textcolor{red}{3.30} & \textcolor{red}{3.50} & \textcolor{red}{0.38} & \textcolor{red}{3.20} \\ \hline
Conv2.5Db & \textcolor{red}{3.33} & \textcolor{red}{0.43} & \textcolor{red}{2.81} & \textcolor{red}{3.72} & \textcolor{red}{0.73} & \textcolor{red}{0.53} \\ \hline
\end{tabular}
\label{TestUF2}
\end{center}
\end{table}

%% file: Tables/testvf2.tex
\begin{table}[H]
\caption{Test error on full velocity}\begin{center}
\begin{tabular}{|c|c|c|c|c|c|c|}
\hline 
& Basic & BN & E & SL & BN \& SL & E \& SL \\ \hline
FC t & \textcolor{red}{1.60} & \textcolor{red}{0.43} & \textcolor{red}{1.61} & \textcolor{red}{1.55} & \textcolor{red}{0.48} & \textcolor{red}{1.55} \\ \hline
Conv2D & \textcolor{red}{3.02} & \textcolor{blue}{0.18} & \textcolor{red}{2.88} & \textcolor{red}{2.95} & \textcolor{blue}{0.24} & \textcolor{red}{3.19} \\ \hline
Conv2D t & \textcolor{red}{20.56} & \textcolor{blue}{0.25} & \textcolor{red}{20.55} & \textcolor{red}{20.55} & \textcolor{blue}{0.26} & \textcolor{red}{20.57} \\ \hline
Conv3D & \textcolor{red}{4.93} & \textbf{0.17} & \textcolor{red}{5.02} & \textcolor{red}{5.02} & \textcolor{blue}{0.23} & \textcolor{red}{5.14} \\ \hline
Conv2.5D & \textcolor{red}{3.22} & \textcolor{blue}{0.24} & \textcolor{red}{3.01} & \textcolor{red}{3.29} & \textcolor{red}{0.43} & \textcolor{red}{3.14} \\ \hline
Conv2.5Db & \textcolor{red}{2.92} & \textcolor{red}{0.58} & \textcolor{red}{2.60} & \textcolor{red}{3.30} & \textcolor{red}{0.54} & \textcolor{red}{0.48} \\ \hline
\end{tabular}
\label{TestVF2}
\end{center}
\end{table}

%% file: Tables/trainufz2.tex
\begin{table}[H]
\caption{Train error on displacement after zoom}\begin{center}
\begin{tabular}{|c|c|c|c|c|c|c|}
\hline 
& Basic & BN & E & SL & BN \& SL & E \& SL \\ \hline
FC t & \textcolor{red}{2.01} & \textcolor{red}{0.49} & \textcolor{red}{2.06} & \textcolor{red}{1.83} & \textcolor{red}{0.47} & \textcolor{red}{1.88} \\ \hline
Conv2D & \textcolor{red}{4.75} & \textcolor{blue}{0.13} & \textcolor{red}{4.80} & \textcolor{red}{4.74} & \textcolor{blue}{0.17} & \textcolor{red}{4.61} \\ \hline
Conv2D t & \textcolor{red}{27.45} & \textcolor{blue}{0.25} & \textcolor{red}{27.46} & \textcolor{red}{27.44} & \textbf{0.13} & \textcolor{red}{27.44} \\ \hline
Conv3D & \textcolor{red}{4.61} & \textcolor{blue}{0.22} & \textcolor{red}{4.47} & \textcolor{red}{4.57} & \textcolor{blue}{0.20} & \textcolor{red}{4.46} \\ \hline
Conv2.5D & \textcolor{red}{4.72} & \textcolor{blue}{0.19} & \textcolor{red}{4.94} & \textcolor{red}{4.57} & \textcolor{blue}{0.16} & \textcolor{red}{4.86} \\ \hline
Conv2.5Db & \textcolor{red}{4.69} & \textcolor{red}{0.28} & \textcolor{red}{5.07} & \textcolor{red}{4.43} & \textcolor{red}{0.51} & \textcolor{red}{0.29} \\ \hline
FC t Boundary & \textcolor{red}{1.59} & \textcolor{red}{0.48} & \textcolor{red}{1.64} & \textcolor{red}{1.58} & \textcolor{red}{0.41} & \textcolor{red}{1.48} \\ \hline
Conv1D Boundary & \textcolor{red}{4.62} & \textcolor{blue}{0.19} & \textcolor{red}{4.80} & \textcolor{red}{4.70} & \textcolor{blue}{0.14} & \textcolor{red}{4.84} \\ \hline
Conv1D t Boundary & \textcolor{red}{27.45} & \textcolor{red}{0.30} & \textcolor{red}{27.45} & \textcolor{red}{27.44} & \textcolor{red}{0.38} & \textcolor{red}{27.45} \\ \hline
Conv2D Boundary & \textcolor{red}{11.25} & \textcolor{blue}{0.15} & \textcolor{red}{11.59} & \textcolor{red}{11.22} & \textcolor{blue}{0.21} & \textcolor{red}{11.36} \\ \hline
Conv1.5D Boundary & \textcolor{red}{5.38} & \textcolor{blue}{0.18} & \textcolor{red}{5.57} & \textcolor{red}{5.60} & \textcolor{blue}{0.18} & \textcolor{red}{5.62} \\ \hline
\end{tabular}
\label{TrainUFZ2}
\end{center}
\end{table}

%% file: Tables/trainvfz2.tex
\begin{table}[H]
\caption{Train error on velocity after zoom}\begin{center}
\begin{tabular}{|c|c|c|c|c|c|c|}
\hline 
& Basic & BN & E & SL & BN \& SL & E \& SL \\ \hline
FC t & \textcolor{red}{2.80} & \textcolor{red}{0.65} & \textcolor{red}{2.85} & \textcolor{red}{2.64} & \textcolor{red}{0.65} & \textcolor{red}{2.71} \\ \hline
Conv2D & \textcolor{red}{4.42} & \textcolor{blue}{0.12} & \textcolor{red}{4.47} & \textcolor{red}{4.41} & \textcolor{blue}{0.15} & \textcolor{red}{4.29} \\ \hline
Conv2D t & \textcolor{red}{22.49} & \textcolor{blue}{0.18} & \textcolor{red}{22.49} & \textcolor{red}{22.48} & \textbf{0.09} & \textcolor{red}{22.47} \\ \hline
Conv3D & \textcolor{red}{4.57} & \textcolor{blue}{0.18} & \textcolor{red}{4.44} & \textcolor{red}{4.52} & \textcolor{blue}{0.18} & \textcolor{red}{4.44} \\ \hline
Conv2.5D & \textcolor{red}{4.45} & \textcolor{red}{0.21} & \textcolor{red}{4.66} & \textcolor{red}{4.35} & \textcolor{blue}{0.17} & \textcolor{red}{4.56} \\ \hline
Conv2.5Db & \textcolor{red}{4.42} & \textcolor{red}{0.39} & \textcolor{red}{4.66} & \textcolor{red}{4.23} & \textcolor{red}{0.55} & \textcolor{red}{0.37} \\ \hline
FC t Boundary & \textcolor{red}{2.35} & \textcolor{red}{0.62} & \textcolor{red}{2.42} & \textcolor{red}{2.28} & \textcolor{red}{0.54} & \textcolor{red}{2.21} \\ \hline
Conv1D Boundary & \textcolor{red}{4.30} & \textcolor{blue}{0.16} & \textcolor{red}{4.45} & \textcolor{red}{4.37} & \textcolor{blue}{0.13} & \textcolor{red}{4.48} \\ \hline
Conv1D t Boundary & \textcolor{red}{22.49} & \textcolor{red}{0.21} & \textcolor{red}{22.49} & \textcolor{red}{22.48} & \textcolor{red}{0.24} & \textcolor{red}{22.49} \\ \hline
Conv2D Boundary & \textcolor{red}{9.58} & \textcolor{blue}{0.11} & \textcolor{red}{9.85} & \textcolor{red}{9.52} & \textcolor{blue}{0.17} & \textcolor{red}{9.62} \\ \hline
Conv1.5D Boundary & \textcolor{red}{5.80} & \textcolor{red}{0.20} & \textcolor{red}{5.93} & \textcolor{red}{5.95} & \textcolor{blue}{0.18} & \textcolor{red}{5.83} \\ \hline
\end{tabular}
\label{TrainVFZ2}
\end{center}
\end{table}

%% file: Tables/testufz2.tex
\begin{table}[H]
\caption{Test error on displacement after zoom}\begin{center}
\begin{tabular}{|c|c|c|c|c|c|c|}
\hline 
& Basic & BN & E & SL & BN \& SL & E \& SL \\ \hline
FC t & \textcolor{red}{1.34} & \textcolor{red}{0.37} & \textcolor{red}{1.37} & \textcolor{red}{1.24} & \textcolor{red}{0.46} & \textcolor{red}{1.29} \\ \hline
Conv2D & \textcolor{red}{3.25} & \textcolor{red}{0.24} & \textcolor{red}{3.15} & \textcolor{red}{3.22} & \textcolor{red}{0.31} & \textcolor{red}{3.46} \\ \hline
Conv2D t & \textcolor{red}{24.45} & \textcolor{red}{0.28} & \textcolor{red}{24.46} & \textcolor{red}{24.47} & \textcolor{red}{0.29} & \textcolor{red}{24.47} \\ \hline
Conv3D & \textcolor{red}{4.72} & \textcolor{red}{0.18} & \textcolor{red}{4.66} & \textcolor{red}{4.64} & \textcolor{red}{0.26} & \textcolor{red}{4.81} \\ \hline
Conv2.5D & \textcolor{red}{3.47} & \textcolor{red}{0.28} & \textcolor{red}{3.30} & \textcolor{red}{3.50} & \textcolor{red}{0.38} & \textcolor{red}{3.20} \\ \hline
Conv2.5Db & \textcolor{red}{3.33} & \textcolor{red}{0.43} & \textcolor{red}{2.81} & \textcolor{red}{3.72} & \textcolor{red}{0.73} & \textcolor{red}{0.53} \\ \hline
FC t Boundary & \textcolor{red}{1.02} & \textcolor{red}{0.22} & \textcolor{red}{1.08} & \textcolor{red}{1.06} & \textcolor{red}{0.24} & \textcolor{red}{1.07} \\ \hline
Conv1D Boundary & \textcolor{red}{5.01} & \textcolor{blue}{0.13} & \textcolor{red}{4.84} & \textcolor{red}{4.94} & \textbf{0.08} & \textcolor{red}{4.79} \\ \hline
Conv1D t Boundary & \textcolor{red}{25.27} & \textcolor{red}{0.26} & \textcolor{red}{25.28} & \textcolor{red}{25.27} & \textcolor{red}{0.36} & \textcolor{red}{25.28} \\ \hline
Conv2D Boundary & \textcolor{red}{13.28} & \textcolor{blue}{0.12} & \textcolor{red}{13.17} & \textcolor{red}{13.04} & \textcolor{blue}{0.13} & \textcolor{red}{13.21} \\ \hline
Conv1.5D Boundary & \textcolor{red}{5.99} & \textcolor{red}{0.29} & \textcolor{red}{6.19} & \textcolor{red}{6.16} & \textcolor{blue}{0.14} & \textcolor{red}{6.03} \\ \hline
\end{tabular}
\label{TestUFZ2}
\end{center}
\end{table}

%% file: Tables/testvfz2.tex
\begin{table}[H]
\caption{Test error on velocity after zoom}\begin{center}
\begin{tabular}{|c|c|c|c|c|c|c|}
\hline 
& Basic & BN & E & SL & BN \& SL & E \& SL \\ \hline
FC t & \textcolor{red}{2.13} & \textcolor{red}{0.43} & \textcolor{red}{2.28} & \textcolor{red}{2.11} & \textcolor{red}{0.43} & \textcolor{red}{2.08} \\ \hline
Conv2D & \textcolor{red}{2.84} & \textcolor{red}{0.14} & \textcolor{red}{2.80} & \textcolor{red}{2.90} & \textcolor{red}{0.19} & \textcolor{red}{3.02} \\ \hline
Conv2D t & \textcolor{red}{21.76} & \textcolor{blue}{0.09} & \textcolor{red}{21.77} & \textcolor{red}{21.76} & \textcolor{blue}{0.09} & \textcolor{red}{21.74} \\ \hline
Conv3D & \textcolor{red}{3.90} & \textcolor{blue}{0.13} & \textcolor{red}{3.98} & \textcolor{red}{3.99} & \textcolor{blue}{0.10} & \textcolor{red}{4.02} \\ \hline
Conv2.5D & \textcolor{red}{3.00} & \textcolor{red}{0.17} & \textcolor{red}{2.97} & \textcolor{red}{3.06} & \textcolor{blue}{0.13} & \textcolor{red}{3.11} \\ \hline
Conv2.5Db & \textcolor{red}{2.97} & \textcolor{red}{0.30} & \textcolor{red}{2.77} & \textcolor{red}{3.26} & \textcolor{red}{0.43} & \textcolor{red}{0.39} \\ \hline
FC t Boundary & \textcolor{red}{1.77} & \textcolor{red}{0.35} & \textcolor{red}{1.76} & \textcolor{red}{1.65} & \textcolor{red}{0.38} & \textcolor{red}{1.79} \\ \hline
Conv1D Boundary & \textcolor{red}{4.39} & \textcolor{blue}{0.12} & \textcolor{red}{4.25} & \textcolor{red}{4.27} & \textbf{0.07} & \textcolor{red}{4.20} \\ \hline
Conv1D t Boundary & \textcolor{red}{21.76} & \textcolor{red}{0.16} & \textcolor{red}{21.77} & \textcolor{red}{21.76} & \textcolor{red}{0.24} & \textcolor{red}{21.76} \\ \hline
Conv2D Boundary & \textcolor{red}{11.51} & \textcolor{blue}{0.12} & \textcolor{red}{11.38} & \textcolor{red}{11.26} & \textcolor{blue}{0.12} & \textcolor{red}{11.40} \\ \hline
Conv1.5D Boundary & \textcolor{red}{6.29} & \textcolor{red}{0.29} & \textcolor{red}{6.38} & \textcolor{red}{6.46} & \textcolor{red}{0.15} & \textcolor{red}{6.42} \\ \hline
\end{tabular}
\label{TestVFZ2}
\end{center}
\end{table}

%% file: Tables/trainepoch.tex
\begin{table}[H]
\caption{Average time of one epoch}\begin{center}
\begin{tabular}{|c|c|c|c|c|c|c|}
\hline 
& Basic & BN & E & SL & BN \& SL & E \& SL \\ \hline
FC t & \textcolor{red}{0.49} & \textcolor{red}{0.49} & \textcolor{red}{0.53} & \textcolor{red}{0.42} & \textcolor{red}{0.46} & \textcolor{red}{0.51} \\ \hline
Conv2D & \textcolor{red}{0.60} & \textcolor{red}{0.63} & \textcolor{red}{0.72} & \textcolor{red}{0.43} & \textcolor{red}{0.44} & \textcolor{red}{0.53} \\ \hline
Conv2D t & \textcolor{blue}{0.37} & \textcolor{red}{0.40} & \textcolor{red}{0.65} & \textcolor{blue}{0.34} & \textcolor{red}{0.37} & \textcolor{red}{0.62} \\ \hline
Conv3D & \textcolor{blue}{0.32} & \textcolor{blue}{0.36} & \textcolor{red}{0.78} & \textcolor{blue}{0.26} & \textcolor{blue}{0.30} & \textcolor{red}{0.73} \\ \hline
Conv2.5D & \textcolor{red}{0.69} & \textcolor{red}{0.75} & \textcolor{red}{1.07} & \textcolor{red}{0.68} & \textcolor{red}{0.75} & \textcolor{red}{1.03} \\ \hline
Conv2.5Db & \textcolor{red}{1.10} & \textcolor{red}{1.21} & \textcolor{red}{1.67} & \textcolor{red}{1.05} & \textcolor{red}{1.08} & \textcolor{red}{1.53} \\ \hline
FC t Boundary & \textcolor{blue}{0.20} & \textcolor{blue}{0.21} & \textcolor{blue}{0.26} & \textbf{0.19} & \textcolor{blue}{0.20} & \textcolor{blue}{0.25} \\ \hline
Conv1D Boundary & \textcolor{blue}{0.35} & \textcolor{blue}{0.36} & \textcolor{red}{0.46} & \textcolor{blue}{0.29} & \textcolor{blue}{0.30} & \textcolor{red}{0.40} \\ \hline
Conv1D t Boundary & \textcolor{blue}{0.34} & \textcolor{blue}{0.36} & \textcolor{red}{0.51} & \textcolor{blue}{0.35} & \textcolor{blue}{0.36} & \textcolor{red}{0.54} \\ \hline
Conv2D Boundary & \textcolor{blue}{0.22} & \textcolor{blue}{0.23} & \textcolor{blue}{0.30} & \textcolor{blue}{0.19} & \textcolor{blue}{0.21} & \textcolor{blue}{0.28} \\ \hline
Conv1.5D Boundary & \textcolor{red}{0.69} & \textcolor{red}{0.73} & \textcolor{red}{1.05} & \textcolor{red}{0.65} & \textcolor{red}{0.69} & \textcolor{red}{1.03} \\ \hline
\end{tabular}
\label{trainepoch}
\end{center}
\end{table}

%% file: Tables/trainuf.tex
\begin{table}[H]
\caption{Train error on full displacement}\begin{center}
\begin{tabular}{|c|c|c|c|c|c|c|}
\hline 
& Basic & BN & E & SL & BN \& SL & E \& SL \\ \hline
FC t & \textcolor{red}{0.88} & \textcolor{red}{0.21} & \textcolor{red}{0.86} & \textcolor{red}{0.76} & \textcolor{red}{0.23} & \textcolor{red}{0.75} \\ \hline
Conv2D & \textcolor{blue}{0.11} & \textcolor{blue}{0.10} & \textcolor{blue}{0.13} & \textcolor{blue}{0.12} & \textbf{0.08} & \textcolor{blue}{0.13} \\ \hline
Conv2D t & \textcolor{red}{0.22} & \textcolor{red}{0.19} & \textcolor{red}{0.24} & \textcolor{red}{0.25} & \textcolor{red}{0.21} & \textcolor{red}{0.28} \\ \hline
Conv3D & \textcolor{blue}{0.12} & \textcolor{blue}{0.09} & \textcolor{blue}{0.12} & \textcolor{blue}{0.14} & \textcolor{blue}{0.11} & \textcolor{blue}{0.15} \\ \hline
Conv2.5D & \textcolor{blue}{0.15} & \textcolor{blue}{0.15} & \textcolor{blue}{0.14} & \textcolor{blue}{0.13} & \textcolor{blue}{0.12} & \textcolor{red}{0.17} \\ \hline
Conv2.5Db & \textcolor{blue}{0.15} & \textcolor{blue}{0.16} & \textcolor{blue}{0.16} & \textcolor{blue}{0.16} & \textcolor{blue}{0.16} & \textcolor{red}{0.19} \\ \hline
\end{tabular}
\label{TrainUF}
\end{center}
\end{table}

%% file: Tables/trainvf.tex
\begin{table}[H]
\caption{Train error on full velocity}\begin{center}
\begin{tabular}{|c|c|c|c|c|c|c|}
\hline 
& Basic & BN & E & SL & BN \& SL & E \& SL \\ \hline
FC t & \textcolor{red}{0.82} & \textcolor{red}{0.28} & \textcolor{red}{0.83} & \textcolor{red}{0.83} & \textcolor{red}{0.24} & \textcolor{red}{0.80} \\ \hline
Conv2D & \textcolor{blue}{0.12} & \textbf{0.08} & \textcolor{blue}{0.12} & \textcolor{blue}{0.11} & \textcolor{blue}{0.08} & \textcolor{blue}{0.11} \\ \hline
Conv2D t & \textcolor{red}{0.25} & \textcolor{red}{0.19} & \textcolor{red}{0.20} & \textcolor{red}{0.23} & \textcolor{red}{0.22} & \textcolor{red}{0.22} \\ \hline
Conv3D & \textcolor{blue}{0.14} & \textcolor{blue}{0.10} & \textcolor{blue}{0.13} & \textcolor{blue}{0.13} & \textcolor{blue}{0.12} & \textcolor{red}{0.15} \\ \hline
Conv2.5D & \textcolor{blue}{0.14} & \textcolor{red}{0.16} & \textcolor{blue}{0.13} & \textcolor{blue}{0.13} & \textcolor{blue}{0.13} & \textcolor{red}{0.16} \\ \hline
Conv2.5Db & \textcolor{red}{0.15} & \textcolor{red}{0.17} & \textcolor{blue}{0.12} & \textcolor{red}{0.16} & \textcolor{red}{0.18} & \textcolor{red}{0.18} \\ \hline
\end{tabular}
\label{TrainVF}
\end{center}
\end{table}

%% file: Tables/testuf.tex
\begin{table}[H]
\caption{Test error on full displacement}\begin{center}
\begin{tabular}{|c|c|c|c|c|c|c|}
\hline 
& Basic & BN & E & SL & BN \& SL & E \& SL \\ \hline
FC t & \textcolor{red}{0.92} & \textcolor{blue}{0.27} & \textcolor{red}{0.90} & \textcolor{red}{0.80} & \textcolor{blue}{0.28} & \textcolor{red}{0.78} \\ \hline
Conv2D & \textcolor{red}{1.12} & \textcolor{red}{0.77} & \textcolor{red}{1.22} & \textcolor{red}{1.13} & \textcolor{red}{0.79} & \textcolor{red}{1.11} \\ \hline
Conv2D t & \textcolor{blue}{0.24} & \textcolor{blue}{0.22} & \textcolor{blue}{0.26} & \textcolor{blue}{0.27} & \textcolor{blue}{0.23} & \textcolor{blue}{0.31} \\ \hline
Conv3D & \textcolor{blue}{0.23} & \textcolor{blue}{0.20} & \textcolor{blue}{0.24} & \textcolor{blue}{0.23} & \textbf{0.19} & \textcolor{blue}{0.24} \\ \hline
Conv2.5D & \textcolor{red}{0.55} & \textcolor{red}{0.64} & \textcolor{red}{0.54} & \textcolor{red}{0.53} & \textcolor{red}{0.58} & \textcolor{red}{0.51} \\ \hline
Conv2.5Db & \textcolor{red}{0.66} & \textcolor{red}{0.63} & \textcolor{red}{0.63} & \textcolor{red}{0.54} & \textcolor{red}{0.68} & \textcolor{red}{0.68} \\ \hline
\end{tabular}
\label{TestUF}
\end{center}
\end{table}

%% file: Tables/testvf.tex
\begin{table}[H]
\caption{Test error on full velocity}\begin{center}
\begin{tabular}{|c|c|c|c|c|c|c|}
\hline 
& Basic & BN & E & SL & BN \& SL & E \& SL \\ \hline
FC t & \textcolor{red}{0.86} & \textcolor{blue}{0.34} & \textcolor{red}{0.86} & \textcolor{red}{0.87} & \textcolor{blue}{0.30} & \textcolor{red}{0.82} \\ \hline
Conv2D & \textcolor{red}{1.12} & \textcolor{red}{0.81} & \textcolor{red}{1.24} & \textcolor{red}{1.09} & \textcolor{red}{0.74} & \textcolor{red}{1.11} \\ \hline
Conv2D t & \textcolor{blue}{0.26} & \textcolor{blue}{0.21} & \textcolor{blue}{0.22} & \textcolor{blue}{0.25} & \textcolor{blue}{0.23} & \textcolor{blue}{0.24} \\ \hline
Conv3D & \textcolor{blue}{0.28} & \textcolor{blue}{0.21} & \textcolor{blue}{0.32} & \textcolor{blue}{0.22} & \textbf{0.20} & \textcolor{blue}{0.24} \\ \hline
Conv2.5D & \textcolor{red}{0.54} & \textcolor{red}{0.63} & \textcolor{red}{0.50} & \textcolor{red}{0.49} & \textcolor{red}{0.55} & \textcolor{red}{0.48} \\ \hline
Conv2.5Db & \textcolor{red}{0.70} & \textcolor{red}{0.72} & \textcolor{red}{0.76} & \textcolor{red}{0.50} & \textcolor{red}{0.64} & \textcolor{red}{0.64} \\ \hline
\end{tabular}
\label{TestVF}
\end{center}
\end{table}

%% file: Tables/trainufz.tex
\begin{table}[H]
\caption{Train error on displacement after zoom}\begin{center}
\begin{tabular}{|c|c|c|c|c|c|c|}
\hline 
& Basic & BN & E & SL & BN \& SL & E \& SL \\ \hline
FC t & \textcolor{red}{0.79} & \textcolor{red}{0.17} & \textcolor{red}{0.75} & \textcolor{red}{0.64} & \textcolor{red}{0.19} & \textcolor{red}{0.61} \\ \hline
Conv2D & \textcolor{red}{0.04} & \textcolor{blue}{0.03} & \textcolor{red}{0.05} & \textcolor{red}{0.05} & \textcolor{blue}{0.03} & \textcolor{red}{0.05} \\ \hline
Conv2D t & \textcolor{red}{0.12} & \textcolor{red}{0.10} & \textcolor{red}{0.12} & \textcolor{red}{0.13} & \textcolor{red}{0.11} & \textcolor{red}{0.16} \\ \hline
Conv3D & \textcolor{red}{0.06} & \textcolor{red}{0.04} & \textcolor{red}{0.05} & \textcolor{red}{0.06} & \textcolor{red}{0.06} & \textcolor{red}{0.08} \\ \hline
Conv2.5D & \textcolor{red}{0.07} & \textcolor{red}{0.08} & \textcolor{red}{0.05} & \textcolor{red}{0.07} & \textcolor{red}{0.06} & \textcolor{red}{0.09} \\ \hline
Conv2.5Db & \textcolor{red}{0.07} & \textcolor{red}{0.08} & \textcolor{red}{0.07} & \textcolor{red}{0.08} & \textcolor{red}{0.07} & \textcolor{red}{0.12} \\ \hline
FC t Boundary & \textcolor{red}{0.55} & \textcolor{red}{0.17} & \textcolor{red}{0.56} & \textcolor{red}{0.52} & \textcolor{red}{0.17} & \textcolor{red}{0.51} \\ \hline
Conv1D Boundary & \textcolor{blue}{0.03} & \textbf{0.02} & \textcolor{blue}{0.03} & \textcolor{blue}{0.04} & \textcolor{blue}{0.02} & \textcolor{red}{0.05} \\ \hline
Conv1D t Boundary & \textcolor{red}{0.09} & \textcolor{red}{0.08} & \textcolor{red}{0.07} & \textcolor{red}{0.12} & \textcolor{red}{0.11} & \textcolor{red}{0.11} \\ \hline
Conv2D Boundary & \textcolor{blue}{0.04} & \textcolor{red}{0.06} & \textcolor{red}{0.06} & \textcolor{red}{0.06} & \textcolor{red}{0.06} & \textcolor{red}{0.05} \\ \hline
Conv1.5D Boundary & \textcolor{red}{0.05} & \textcolor{blue}{0.02} & \textcolor{red}{0.04} & \textcolor{red}{0.05} & \textcolor{blue}{0.03} & \textcolor{blue}{0.04} \\ \hline
\end{tabular}
\label{TrainUFZ}
\end{center}
\end{table}

%% file: Tables/trainvfz.tex
\begin{table}[H]
\caption{Train error on velocity after zoom}\begin{center}
\begin{tabular}{|c|c|c|c|c|c|c|}
\hline 
& Basic & BN & E & SL & BN \& SL & E \& SL \\ \hline
FC t & \textcolor{red}{1.23} & \textcolor{red}{0.24} & \textcolor{red}{1.18} & \textcolor{red}{1.05} & \textcolor{red}{0.27} & \textcolor{red}{0.99} \\ \hline
Conv2D & \textcolor{red}{0.05} & \textcolor{blue}{0.04} & \textcolor{red}{0.06} & \textcolor{red}{0.05} & \textcolor{blue}{0.03} & \textcolor{red}{0.06} \\ \hline
Conv2D t & \textcolor{red}{0.12} & \textcolor{red}{0.12} & \textcolor{red}{0.12} & \textcolor{red}{0.16} & \textcolor{red}{0.13} & \textcolor{red}{0.17} \\ \hline
Conv3D & \textcolor{red}{0.07} & \textcolor{red}{0.05} & \textcolor{red}{0.06} & \textcolor{red}{0.07} & \textcolor{red}{0.06} & \textcolor{red}{0.09} \\ \hline
Conv2.5D & \textcolor{red}{0.08} & \textcolor{red}{0.07} & \textcolor{red}{0.07} & \textcolor{red}{0.09} & \textcolor{red}{0.07} & \textcolor{red}{0.10} \\ \hline
Conv2.5Db & \textcolor{red}{0.09} & \textcolor{red}{0.10} & \textcolor{red}{0.10} & \textcolor{red}{0.11} & \textcolor{red}{0.10} & \textcolor{red}{0.12} \\ \hline
FC t Boundary & \textcolor{red}{0.90} & \textcolor{red}{0.25} & \textcolor{red}{0.92} & \textcolor{red}{0.84} & \textcolor{red}{0.25} & \textcolor{red}{0.84} \\ \hline
Conv1D Boundary & \textcolor{blue}{0.03} & \textbf{0.02} & \textcolor{blue}{0.03} & \textcolor{blue}{0.03} & \textcolor{blue}{0.02} & \textcolor{blue}{0.04} \\ \hline
Conv1D t Boundary & \textcolor{red}{0.09} & \textcolor{red}{0.09} & \textcolor{red}{0.08} & \textcolor{red}{0.11} & \textcolor{red}{0.11} & \textcolor{red}{0.12} \\ \hline
Conv2D Boundary & \textcolor{blue}{0.04} & \textcolor{red}{0.05} & \textcolor{red}{0.06} & \textcolor{red}{0.06} & \textcolor{red}{0.07} & \textcolor{red}{0.05} \\ \hline
Conv1.5D Boundary & \textcolor{red}{0.05} & \textcolor{blue}{0.03} & \textcolor{red}{0.05} & \textcolor{red}{0.05} & \textcolor{blue}{0.04} & \textcolor{red}{0.05} \\ \hline
\end{tabular}
\label{TrainVFZ}
\end{center}
\end{table}

%% file: Tables/testufz.tex
\begin{table}[H]
\caption{Test error on displacement after zoom}\begin{center}
\begin{tabular}{|c|c|c|c|c|c|c|}
\hline 
& Basic & BN & E & SL & BN \& SL & E \& SL \\ \hline
FC t & \textcolor{red}{0.92} & \textcolor{red}{0.27} & \textcolor{red}{0.90} & \textcolor{red}{0.80} & \textcolor{red}{0.28} & \textcolor{red}{0.78} \\ \hline
Conv2D & \textcolor{red}{1.12} & \textcolor{red}{0.77} & \textcolor{red}{1.22} & \textcolor{red}{1.13} & \textcolor{red}{0.79} & \textcolor{red}{1.11} \\ \hline
Conv2D t & \textcolor{red}{0.24} & \textcolor{blue}{0.22} & \textcolor{red}{0.26} & \textcolor{red}{0.27} & \textcolor{red}{0.23} & \textcolor{red}{0.31} \\ \hline
Conv3D & \textcolor{red}{0.23} & \textcolor{blue}{0.20} & \textcolor{red}{0.24} & \textcolor{red}{0.23} & \textcolor{blue}{0.19} & \textcolor{red}{0.24} \\ \hline
Conv2.5D & \textcolor{red}{0.55} & \textcolor{red}{0.64} & \textcolor{red}{0.54} & \textcolor{red}{0.53} & \textcolor{red}{0.58} & \textcolor{red}{0.51} \\ \hline
Conv2.5Db & \textcolor{red}{0.66} & \textcolor{red}{0.63} & \textcolor{red}{0.63} & \textcolor{red}{0.54} & \textcolor{red}{0.68} & \textcolor{red}{0.68} \\ \hline
FC t Boundary & \textcolor{red}{0.60} & \textcolor{red}{0.23} & \textcolor{red}{0.62} & \textcolor{red}{0.58} & \textcolor{red}{0.25} & \textcolor{red}{0.58} \\ \hline
Conv1D Boundary & \textcolor{red}{1.51} & \textcolor{red}{1.42} & \textcolor{red}{1.55} & \textcolor{red}{1.57} & \textcolor{red}{1.33} & \textcolor{red}{1.45} \\ \hline
Conv1D t Boundary & \textcolor{blue}{0.13} & \textcolor{blue}{0.11} & \textbf{0.11} & \textcolor{blue}{0.16} & \textcolor{blue}{0.14} & \textcolor{blue}{0.15} \\ \hline
Conv2D Boundary & \textcolor{red}{0.30} & \textcolor{red}{0.24} & \textcolor{red}{0.28} & \textcolor{red}{0.25} & \textcolor{red}{0.23} & \textcolor{red}{0.25} \\ \hline
Conv1.5D Boundary & \textcolor{red}{0.78} & \textcolor{red}{1.07} & \textcolor{red}{1.03} & \textcolor{red}{0.96} & \textcolor{red}{0.85} & \textcolor{red}{1.02} \\ \hline
\end{tabular}
\label{TestUFZ}
\end{center}
\end{table}

%% file: Tables/testvfz.tex
\begin{table}[H]
\caption{Test error on velocity after zoom}\begin{center}
\begin{tabular}{|c|c|c|c|c|c|c|}
\hline 
& Basic & BN & E & SL & BN \& SL & E \& SL \\ \hline
FC t & \textcolor{red}{1.25} & \textcolor{red}{0.30} & \textcolor{red}{1.21} & \textcolor{red}{1.08} & \textcolor{red}{0.33} & \textcolor{red}{1.01} \\ \hline
Conv2D & \textcolor{red}{1.01} & \textcolor{red}{0.85} & \textcolor{red}{1.08} & \textcolor{red}{0.95} & \textcolor{red}{0.88} & \textcolor{red}{1.09} \\ \hline
Conv2D t & \textcolor{blue}{0.16} & \textcolor{blue}{0.16} & \textcolor{blue}{0.15} & \textcolor{blue}{0.19} & \textcolor{blue}{0.17} & \textcolor{blue}{0.20} \\ \hline
Conv3D & \textcolor{blue}{0.18} & \textcolor{blue}{0.18} & \textcolor{blue}{0.19} & \textcolor{blue}{0.19} & \textcolor{blue}{0.16} & \textcolor{blue}{0.20} \\ \hline
Conv2.5D & \textcolor{red}{0.61} & \textcolor{red}{0.76} & \textcolor{red}{0.58} & \textcolor{red}{0.60} & \textcolor{red}{0.69} & \textcolor{red}{0.54} \\ \hline
Conv2.5Db & \textcolor{red}{0.76} & \textcolor{red}{0.77} & \textcolor{red}{0.74} & \textcolor{red}{0.61} & \textcolor{red}{0.84} & \textcolor{red}{0.82} \\ \hline
FC t Boundary & \textcolor{red}{0.93} & \textcolor{red}{0.31} & \textcolor{red}{0.96} & \textcolor{red}{0.88} & \textcolor{red}{0.33} & \textcolor{red}{0.88} \\ \hline
Conv1D Boundary & \textcolor{red}{1.42} & \textcolor{red}{1.35} & \textcolor{red}{1.46} & \textcolor{red}{1.48} & \textcolor{red}{1.26} & \textcolor{red}{1.36} \\ \hline
Conv1D t Boundary & \textcolor{blue}{0.14} & \textcolor{blue}{0.13} & \textbf{0.12} & \textcolor{blue}{0.16} & \textcolor{blue}{0.14} & \textcolor{blue}{0.16} \\ \hline
Conv2D Boundary & \textcolor{red}{0.31} & \textcolor{red}{0.24} & \textcolor{red}{0.29} & \textcolor{red}{0.25} & \textcolor{blue}{0.23} & \textcolor{red}{0.25} \\ \hline
Conv1.5D Boundary & \textcolor{red}{0.75} & \textcolor{red}{1.03} & \textcolor{red}{0.96} & \textcolor{red}{0.90} & \textcolor{red}{0.80} & \textcolor{red}{0.95} \\ \hline
\end{tabular}
\label{TestVFZ}
\end{center}
\end{table}